\documentclass[lettersize,journal]{IEEEtran}
\usepackage{amsmath,amsfonts}
\usepackage{algorithmic}
\usepackage{algorithm}
\usepackage{array}
\usepackage[caption=false,font=normalsize,labelfont=sf,textfont=sf]{subfig}
\usepackage{textcomp}
\usepackage{stfloats}
\usepackage{url}
\usepackage{verbatim}
\usepackage{graphicx}
\usepackage{cite}
\usepackage{graphicx}
\usepackage{amsmath}
\usepackage{amssymb}
\usepackage{booktabs}
\usepackage{color}  

\usepackage{bm}
\usepackage{multirow}
\usepackage{bbding}

\begin{document}

\title{Shadow Removal by High-Quality Shadow Synthesis}

\author{Yunshan Zhong,
Lizhou You,
Yuxin Zhang,
Fei Chao, \\
Yonghong Tian,~\IEEEmembership{Fellow, IEEE},
Rongrong Ji,~\IEEEmembership{Senior Member, IEEE}
\thanks{This work was supported by National Key R\&D Program of China (No.2022ZD0118202), the National Science Fund for Distinguished Young Scholars (No.62025603), the National Natural Science Foundation of China (No. U21B2037, No. U22B2051, No. 62176222, No. 62176223, No. 62176226, No. 62072386, No. 62072387, No. 62072389, No. 62002305 and No. 62272401), and the Natural Science Foundation of Fujian Province of China (No.2021J01002,  No.2022J06001).
}
\thanks{Y. Zhong is with Institute of Artificial Intelligence, Department of Artificial Intelligence, School of Informatics, and Key Laboratory of Multimedia Trusted Perception and Efficient Computing, Ministry of Education of China, Xiamen University, Xiamen 361005, P.R. China.}
\thanks{L. You, Y. Zhang, and C. Fei are with Department of Artificial Intelligence, School of Informatics, and Key Laboratory of Multimedia Trusted Perception and Efficient Computing, Ministry of Education of China, Xiamen University, Xiamen 361005, P.R. China.}
\thanks{Y. Tian is with the Peng Cheng Laboratory, Shenzhen 518066,
China, and also with the National Engineering Laboratory for Video Technology (NELVT), School of Electronics Engineering and Computer Science, Peking University, Beijing 100871, China.}
\thanks{R. Ji (Corresponding Author) is with Institute of Artificial Intelligence, and Key Laboratory of Multimedia Trusted Perception and Efficient Computing, Ministry of Education of China, Xiamen University, Xiamen 361005, P.R. China, and also with the Peng Cheng Laboratory, Shenzhen 518000, P.R. China (e-mail: rrji@xmu.edu.cn).}
\thanks{Manuscript received April 19, 2021; revised August 16, 2021.}}

\markboth{Journal of \LaTeX\ Class Files,~Vol.~14, No.~8, August~2021}%
{Shell \MakeLowercase{\textit{et al.}}: A Sample Article Using IEEEtran.cls for IEEE Journals}

\IEEEpubid{0000--0000/00\$00.00~\copyright~2021 IEEE}

\maketitle

\begin{abstract}
Most shadow removal methods rely on the invasion of training images associated with laborious and lavish shadow region annotations, leading to the increasing popularity of shadow image synthesis.
However, the poor performance also stems from these synthesized images since they are often shadow-inauthentic and details-impaired.
In this paper, we present a novel generation framework, referred to as HQSS, for high-quality pseudo shadow image synthesis.  
The given image is first decoupled into a shadow region identity and a non-shadow region identity.
HQSS employs a shadow feature encoder and a generator to synthesize pseudo images.
Specifically, the encoder extracts the shadow feature of a region identity which is then paired with another region identity to serve as the generator input to synthesize a pseudo image.
The pseudo image is expected to have the shadow feature as its input shadow feature and as well as a real-like image detail as its input region identity.
To fulfill this goal, we design three learning objectives.
When the shadow feature and input region identity are from the same region identity, we propose a self-reconstruction loss that guides the generator to reconstruct an identical pseudo image as its input.
When the shadow feature and input region identity are from different identities, we introduce an inter-reconstruction loss and a cycle-reconstruction loss to make sure that shadow characteristics and detail information can be well retained in the synthesized images. 
Our HQSS is observed to outperform the state-of-the-art methods on ISTD dataset, Video Shadow Removal dataset, and SRD dataset.
The code is available at \url{https://github.com/zysxmu/HQSS}.
\end{abstract}


\section{Introduction}
\IEEEPARstart{S}{hadow} refers to the dark region in a scene where its light from a light source is blocked by an opaque object.
Originating from the complex interactions among light sources, geometric location, and materials of the objects in the scene~\cite{le2019shadow}, the shadow is a common natural phenomenon that is widely spotted in images and videos.
Shadow itself greatly differs in illuminations and shapes. This greatly affects the color and appearance of objects that are of vital importance to many downstream image processing tasks such as image stitching and alignment~\cite{nie2021unsupervised,zhang2020content,shen2020ransac}, texture analysis~\cite{wang2008texture}, image  inpainting~\cite{yu2018generative,liu2018image,yeh2017semantic}, image denoising~\cite{Zamir2020MIRNet,zamir2021multi,zamir2022restormer} and image deblurring~\cite{tao2018scale,zhang2019deep}.
Consequently, great efforts have been made on shadow removal in the past decades~\cite{huang2011characterizes,finlayson2005removal,guo2012paired,khan2015automatic,zhang2015shadow,yang2012shadow}.

\IEEEpubidadjcol

The remarkable success of these progresses mostly relies on the permit of accessing a large-scale training dataset with paired shadow/non-shadow annotations~\cite{wang2018stacked} to allow end-to-end training in modern deep convolutional neural networks~\cite{le2020shadow}. 
However, annotating datasets is a time laborious and lavish process, in particular for shadow removal task~\cite{hu2019mask}.
It is almost impossible to capture shadow/non-shadow pairs that share consistent colors and luminosity in an uncontrolled nature environment, since the camera pose, the position, and the intensity of the light-source change from time to time~\cite{hu2019mask,liu2021shadowCVPR}. As pointed out by~\cite{le2020shadow}, existing shadow removal datasets, to some extent, exhibit color inconsistency between the shadow image and the corresponding non-shadow counterpart.
Thus, the collected image pairs are often not well aligned.
While the mismatch in colors can be avoided in a highly-controlled lab environment, the generalization capability of the trained model will be undermined since the lab environment only provides very limited and constant scenes~\cite{liu2021shadow,gao2022towards}.

\begin{figure*}[!htbp]
\centering
\subfloat[Non-shadow Images]{
\includegraphics[scale=0.75]{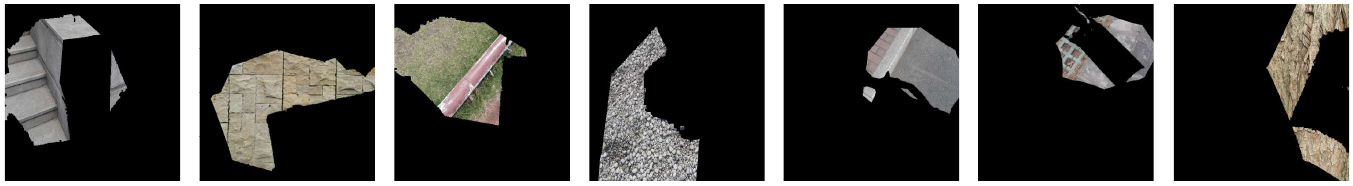} \label{fig:insight-nonshadow}
}
\quad
\subfloat[Real Shadow Images]{
\includegraphics[scale=0.75]{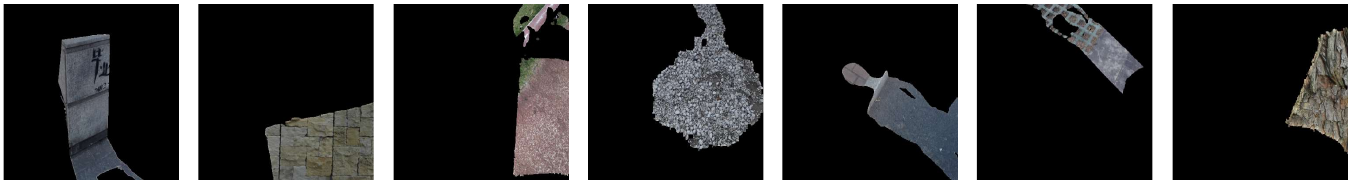} \label{fig:insight-realshadow}
}
\quad
\subfloat[Pseudo Shadow Images by G2R-ShadowNet~\cite{liu2021shadowCVPR}]{
\includegraphics[scale=0.75]{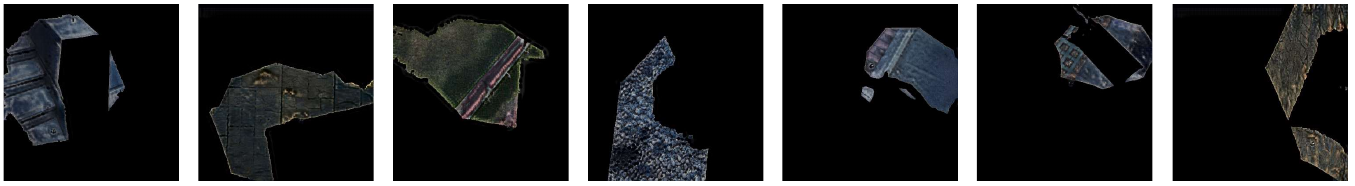} \label{fig:insight-G2R}
}
\quad
\subfloat[Pseudo Shadow Images by our HQSS]{
\includegraphics[scale=0.75]{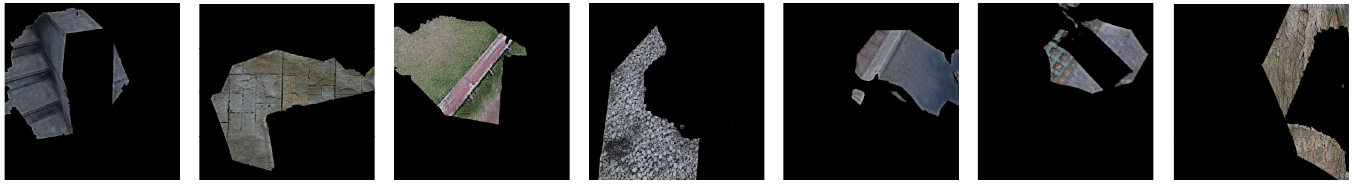} \label{fig:insight-ours}
}
\caption{(a) Real non-shadow images. (b) Real shadow images. (c) Pseudo shadow images by G2R-ShadowNet~\cite{liu2021shadowCVPR}. (d) Pseudo shadow images by our HQSS. Compared with shadow-inauthentic and details-impaired pseudo shadow images by G2R-ShadowNet, ours are more similar to real shadow images and preserve more details of the non-shadow images.}
\label{fig:insight}
\end{figure*}

Consequently, recent attention focuses on performing shadow removal with unpaired training data consisting of shadow images and random non-shadow images~\cite{hu2019mask,liu2021shadow}. Nevertheless, an arbitrarily collected shadow set and non-shadow set often suffer a large domain gap, causing poor performance~\cite{liu2021shadowCVPR,le2020shadow,li2019asymmetric,choi2018stargan}. 
Luckily, a shadow image usually consists of the shadow region and shadow-free regions. Thus, cropping shadow and shadow-free regions from shadow images can well solve the domain gap since they are from the same images. Based on this, one of the popular research lines~\cite{le2020shadow,liu2021shadowCVPR} is to synthesize pseudo shadow images given only shadow images, and then use the pseudo shadow to train the following shadow removal models.
For example, FSS2SR~\cite{le2020shadow} constructs a weakly-supervised framework where the supervisory signal comprises the non-shadow patches directly cropped from the shadow images.
G2R-ShadowNet~\cite{liu2021shadowCVPR} constructs a cycle-GAN~\cite{zhu2017unpaired} based system to produce pseudo shadows for the non-shadow part and train the shadow removal models concurrently.

Nevertheless, the limited performance increase also stems from the poor quality of these synthesized pseudo images since they are often shadow-inauthentic and details-impaired.
Taking images in Fig.\,\ref{fig:insight-G2R} from G2R-ShadowNet~\cite{liu2021shadowCVPR} as an example, we can observe two phenomena.
First, the real shadow images in Fig.\,\ref{fig:insight-realshadow} possess adequate details, just like their non-shadow versions in Fig.\,\ref{fig:insight-nonshadow}. In contrast, details of the pseudo shadow images in Fig.\,\ref{fig:insight-G2R} are either lost or distorted compared with the corresponding non-shadow images. 
Second, the brightness of the pseudo shadow deviates a lot from the real shadow (darker in illustration of Fig.\,\ref{fig:insight-G2R}). Thus, the pseudo images from existing methods are inauthentic. Simply reducing the brightness doesn't mean resembling a real shadow. 
Therefore, further exploration of synthesized pseudo images of high quality is urgent for community development.

\begin{figure*}[!t]
\centering
\includegraphics[width=0.95\textwidth]{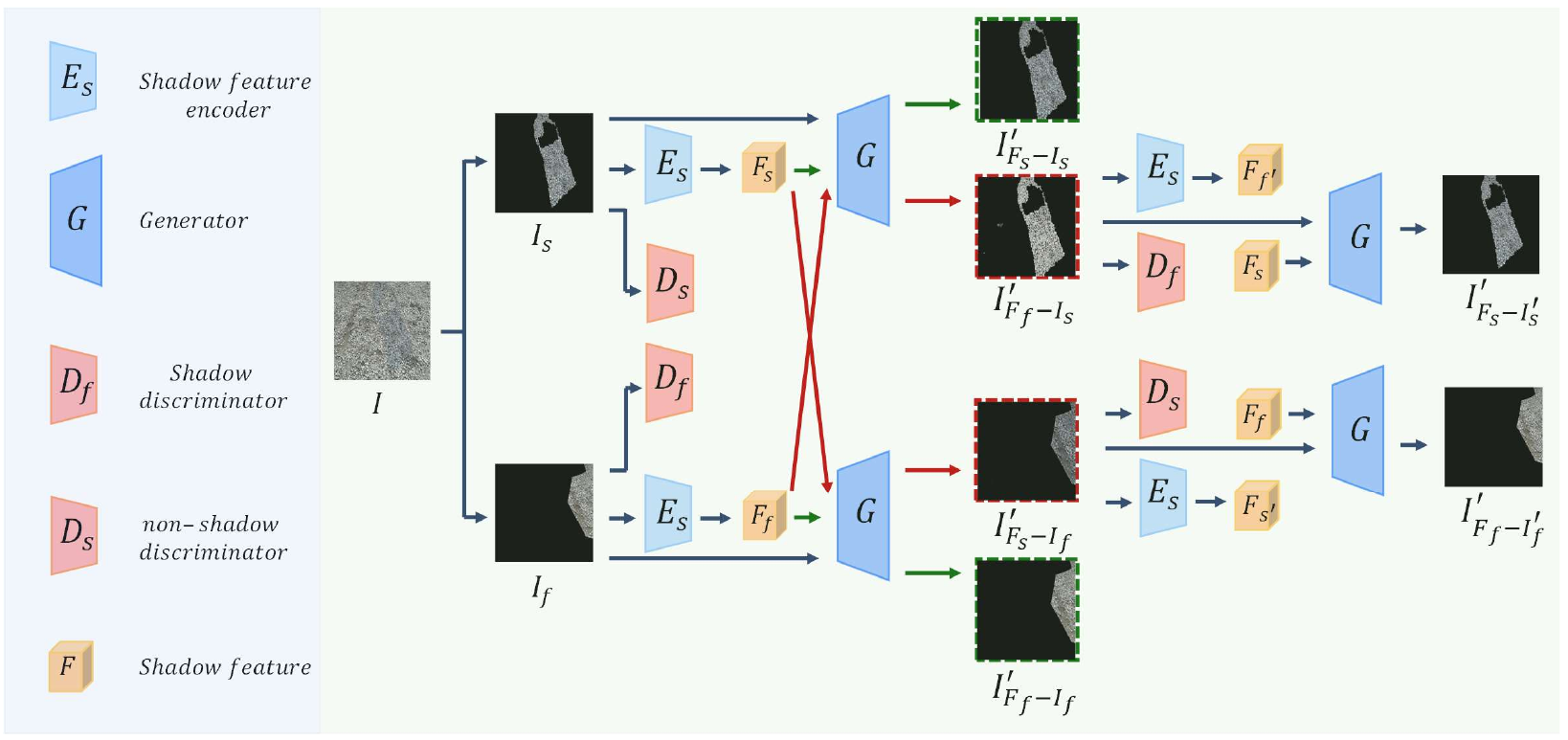}
\caption{An overview of the proposed HQSS. We deploy a shadow feature encoder $E$ and a generator $G$ to synthesize high-quality pseudo images, and a shadow discriminator $D_s$ and a non-shadow discriminator $D_f$ to assist in the training.
For the learning of our HQSS, we design three training objectives including self-reconstruction, inter-reconstruction, and cycle-reconstruction. The images are represented on RGB color space images rather than LAB for easy understanding.
}
\label{fig:framework}
\end{figure*}

In this paper, we focus on the research line of synthesizing pseudo shadow images given only shadow images to train the following shadow removal models.
We present a novel generation framework for high-quality pseudo shadow image synthesis, dubbed HQSS. 
%
%The framework is illustrated in Fig.\,\ref{fig:framework}. As it can be seen, the input image $I$ is first decoupled into a shadow region identity $I_s$ and a non-shadow region identity $I_f$ according to its ground-truth shadow mask and a randomly generated non-shadow mask, respectively.
%
The framework of our HQSS is illustrated in Fig.\,\ref{fig:framework}. Given an input image $I$, we first decouple it into a shadow region identity $I_s$ and a non-shadow region identity $I_f$ according to its ground-truth shadow mask and a randomly generated non-shadow mask.
Two treasures are explored in the decoupled region identity $I_i$ where $i \in \{s, f\}$ includes shadow features such as lightness and detail information such as textures.
To retain these two treasures in the pseudo shadow images, we employ a shadow feature encoder $E$ and a generator $G$. The former is responsible for extracting the shadow feature $F_i$ of the input $I_i$ while the latter synthesize a pseudo image when pairing $I_i$ and $F_j$ where $j \in \{s, f\}$ as its input. The pseudo image is expected to have the shadow feature of the input shadow feature $F_j$, as well as a real-like image detail in the region identify $I_i$. This is achieved by three learning objectives.
The first one, termed self-reconstruction loss detailed in Sec.\,\ref{self-reconstruction}, guides the generator to reconstruct an identical pseudo image as the input when $i = j$.
The remaining two losses, respectively called inter-reconstruction in Sec.\,\ref{inter-reconstruction} and the cycle-reconstruction loss in Sec.\,\ref{cycle-reconstruction}, intend to solve the case of $i$$\neq$$j$ which indicates two different input region identities. These two losses ensure that shadow characteristics and detail information can be retained in the synthesized images.

Combining the novel generation framework and the proposed three learning objectives, our HQSS is demonstrated to afford authentic shadow characteristics and rich detail information as illustrated in Fig.\,\ref{fig:insight-ours}.
When applying the pseudo shadow images to the follow-up shadow removal network, extensive experiments show that our HQSS provides state-of-the-art performance than many existing methods.

\section{Related Work}

In this section, we briefly review the related works on shadow removal and shadow generation.

\subsection{Shadow Removal}

Conventional physical methods resort to some prior knowledge about the shadow properties, such as gradient~\cite{gryka2015learning}, illumination ~\cite{zhang2015shadow,shor2008shadow,xiao2013fast,zhang2018improving,liu2008texture}, and spatial correlation~\cite{guo2012paired,vicente2017leave,fredembach2005hamiltonian}, to calculate pixel values of the shadow region.
Recently, CNN-based approaches trained with large-scale paired datasets become dominant for their significant performance improvements~\cite{liu2021shadow,ding2019argan,wang2018stacked,qu2017deshadownet,hu2019direction,zhang2020ris}.
Qu \emph{et al.}~\cite{qu2017deshadownet} constructed an end-to-end multi-context architecture to complete shadow removal.
Wang \emph{et al.}~\cite{wang2018stacked} introduced a stacked conditional generative adversarial network to perform shadow detection and removal simultaneously.
ARGAN~\cite{ding2019argan} progressively detects and removes shadows in an attentive recurrent generative adversarial network.
Cun \emph{et al.}~\cite{cun2020towards} hierarchically aggregated the dilated multi-context features and attention to detect and remove shadows concurrently.
Auto-Exposure~\cite{fu2021auto} formulates shadow removal as an exposure fusion problem. It first estimates multiple over-exposure images and then fuses the original input with these images to produce the non-shadow images via adaptive learned fusion weights.
In CANet~\cite{chen2021canet}, a two-stage context-aware network is introduced in which the information of the non-shadow patches is transferred into similar shadow patches.

%To avoid the dependence on the paired shadow/non-shadow data, Mask-ShadowGAN~\cite{hu2019mask} leverages unpaired data to learn the bidirectional transformation between shadow/non-shadow image via cycle-consistency constraints~\cite{zhu2017unpaired} and adversarial training.
%
%LG-ShadowNet~\cite{liu2021shadow} exploits the lightness information. They train the first network to learn lightness features and use the learned features to guide the learning of another network for shadow removal.
%
%Recently, some works try to get rid of the non-shadow images to further lower the cost of data collection and control the domain gap between the shadow set and the non-shadow set.
%
%They focus on conducting the shadow removal task by only using the shadow image and its corresponding shadow mask.
%
%In FSS2SR~\cite{le2020shadow}, the original shadow image is cropped into many patches to construct unpaired data which are used to train two networks to predict the shadow parameters and the matte layer, respectively, in an adversarial manner.
%
%G2R-ShadowNet~\cite{liu2021shadowCVPR} instead designs cycle-GAN based system.
%
%The input non-shadow images are first transferred to pseudo shadow images. Then these pseudo shadow image and their non-shadow counterpart form the paired data for training the shadow removal network.

To confront the expensive costs of annotating paired images, Mask-ShadowGAN~\cite{hu2019mask} leverages unpaired data to learn the bidirectional transformation with cycle-consistency constraints~\cite{zhu2017unpaired} and adversarial training.
LG-ShadowNet~\cite{liu2021shadow} first trains a network to extract lightness features, which are then used to guide the learning of another shadow removal network.
To pursue a cheaper data cost and smaller domain bias, some recent methods conduct shadow removal simply using each shadow image and its corresponding shadow mask.
In FSS2SR~\cite{le2020shadow}, the unpaired samples are cropped from the original shadow images and used to adversarially train the shadow parameter prediction network and matte prediction network.
G2R-ShadowNet~\cite{liu2021shadowCVPR} constructs a cycle-GAN based system where the input non-shadow image is first transferred to a pseudo shadow image to form the paired training data for the shadow removal network.
Jin \emph{et al.}~\cite{jin2021dc} integrate shadow and shadow-free domain classifiers into the shadow removal model and introduce novel losses based on physics-based shadow-free chromaticity, shadow-robust perceptual features, and boundary smoothness.

\subsection{Shadow Generation}

%
%Shadow generation draws increasing attention due to the high cost of collecting large-scale paired shadow/non-shadow datasets.
%
%One way to generate shadows is employing inverse rendering which requires information about geometry, illumination, reflectance, and material properties~\cite{sato2003illumination,arief2012realtime,karsch2011rendering,karsch2014automatic,gardner2019deep,zhang2019all,hold2019deep}.
%
%Another way is applying the generative adversarial network (GAN)~\cite{goodfellow2014generative} where the shadow generation is regarded as a particular style transfer task. 
%
%Zhang \emph{et al.}~\cite{zhang2019shadowgan} adopts GAN to synthesize shadows for virtual objects. To obtain training samples, they employ public 3D models and rendering technology to produce the training dataset.
%
%ARShadowGAN~\cite{liu2020arshadowgan} exploits GAN and attention mechanism to learn the mapping between the virtual object shadow and the real-world environment without illumination and geometry estimation.
%
%To eliminate the reliance on the paired non-shadow images, \cite{hu2019mask} and \cite{liu2021shadow} both propose to produce pseudo shadow via deploying a generator to model the mapping relationship between the shadow images and unpaired non-shadow images.  
%
%In \cite{liu2021shadowCVPR} samples the shadow and non-shadow parts from the original shadow images to further eliminate the dependence on non-shadow images. A cycle-GAN based system is employed to generate pseudo shadows for the training of following shadow removal networks.

The heavy expense of annotating massive paired datasets leads to ever-increasing attention on shadow generation.
The inverse rendering-based methods require information about geometry, illumination, reflectance, and material properties~\cite{sato2003illumination,arief2012realtime,karsch2011rendering,karsch2014automatic,gardner2019deep,zhang2019all,hold2019deep}.
Another group considers generative adversarial network (GAN)~\cite{goodfellow2014generative} where the shadow generation is regarded as a particular style transfer task. 

Zhang \emph{et al.}~\cite{zhang2019shadowgan} adopted GAN to synthesize shadows for virtual objects. To obtain training samples, they employed public 3D models and rendering technology to produce the training dataset.
ARShadowGAN~\cite{liu2020arshadowgan} exploits GAN and attention mechanism to learn the mapping between the virtual object shadow and the real-world environment without illumination and geometry estimation.
To eliminate the reliance on the paired non-shadow images, recent studies~\cite{hu2019mask} and \cite{liu2021shadow} produce pseudo shadow via deploying a generator to model the mapping relationship between the shadow images and unpaired non-shadow images.  
Liu \emph{et al.}~\cite{liu2021shadowCVPR} sampled the shadow and non-shadow parts from the original shadow images to further eliminate the dependence on non-shadow images. A cycle-GAN based system is employed to generate pseudo shadows for the training of following shadow removal networks.

\section{Methodology}

%In this section, we first briefly introduce {\color{blue}the sub-networks} of our proposed HQSG, and then we respectively introduce the training objectives including self-reconstruction, inter-reconstruction, and cycle-reconstruction for producing high-quality pseudo shadow images. Finally, we detail our network training process. 

%\subsection{Proposed Networks}
\subsection{Overall Framework}

%In our method, all the input images and output images are represented in LAB color space.
%
%As illustrated in Fig.\,\ref{fig:framework}, the input to the proposed HQSG is an whole shadow image $I$. The whole shadow image $I$ is then cropped to the shadow region $I_s$ and a non-shadow region $I_f$ according to its shadow mask and a random shadow mask, respectively. With region $I_i, i \in \{s,f\}$, our HQSG employs two sub-networks to synthesize high-quality pseudo images and two discriminators to assist in the training. 

Following most existing research on shadow removal ~\cite{liu2021shadow,liu2021shadowCVPR,hu2019direction}, all images including the inputs and outputs in our HQSS are represented on LAB color space.
As illustrated in Fig.\,\ref{fig:framework}, in our HQSS framework, a given LAB image $I$ is decoupled into a shadow region identity $I_s$ according to its associated ground-truth shadow mask and a non-shadow region identity $I_f$ according to a randomly generated non-shadow mask.
Taking a close look at the decoupled region identity $I_i$ where $i \in \{s, f\}$, we realize that it provides two valuable hints for network learning including the shadow feature and detail information.
Specifically, the shadow feature refers to the shadow characteristics of $I_i$ such as the lightness. Note that the non-shadow identity $I_f$ also possesses shadow peculiarity such as the same lightness as the remaining part of the whole image $I$.
In contrast, the detail information is related to the target texture, edge, \emph{etc}.

Our HQSS aims to produce pseudo images that preserve both shadows and details in $I_i$.
To that effect, as illustrated in Fig.\,\ref{fig:framework}, the network backbone of our HQSS is mainly comprised of a shadow feature encoder $E$ and a generator $G$. The encoder $E$ is responsible for extracting the shadow feature of $I_i$ as:
\begin{equation}
        F_i = E(I_i).
\label{extract}
\end{equation}

Afterward, with the shadow feature $F_j$ where $j \in \{s, f\}$ and a decoupled region $I_i$ as its input, the generator intends to produce a pseudo image $I'_{F_j, I_i}$ as:
\begin{equation}
    \begin{aligned}
        I'_{F_j, I_i} = G \big ( C(F_j, I_i) \big ),
    \end{aligned}
\label{generate}
\end{equation}
where $C(\cdot, \cdot)$ indicates the concatenation operation. 
Then, the synthesized pseudo images will be used to train the follow-up shadow removal network.

For a high-qualified shadow synthesis task, we anticipate the pseudo image $I'_{F_j, I_i}$ to have an in-distribution shadow feature as its input shadow feature $F_{j}$ as well as a real-like image detail as its input detail image $I_i$.
Considering the case of $i = j$ which indicates the same region identity as the inputs of both encoder and generator, we introduce a self-reconstruction constraint to empower the generator $G$ with the capacity to reconstruct the input itself in Sec.\,\ref{self-reconstruction}.
Considering the case of $i \neq j$ which means different region identities of the encoder and generator, we introduce an inter-reconstruction loss in Sec.\,\ref{inter-reconstruction} and a cycle-reconstruction loss in Sec.\,\ref{cycle-reconstruction} to make sure that shadow characteristics and detail information can be well retained in the synthesized images.
More details are discussed in the following.

\subsection{Self-reconstruction}\label{self-reconstruction}

We first detail our self-reconstruction constraint to display how the generator $G$ learns to reconstruct an identical pseudo image as the input.
%
%
%The first training objective is self-reconstruction.
%
%The generator $G$ learns to reconstruct an identical pseudo image as the input region given the reference image and the details image are both the input itself.
%
%
%In particular, the shadow region $I_s$ and non-shadow region $I_f$ are used as the reference images. The shadow feature encoder $E$ is adopted to obtain their shadow feature:
%
Given shadow region $I_s$ and non-shadow region $I_f$, we first feed them to the encoder to derive their shadow features as:
\begin{equation}
        F_s = E(I_s), \quad F_f = E(I_f).
\label{ext-E}
\end{equation}

%Then, $F_s$ and $I_s$ are simultaneously input to the generator $G$ to produce a pseudo shadow image $I'_{F_s-I_s}$. Similarly, $F_f$ and $I_f$ are simultaneously input to the generator $G$ to produce a pseudo non-shadow image $I'_{F_f-I_f}$:
%
%\begin{equation}
%        I'_{F_s-I_s} = G \big ( C(F_s, I_s) \big ), \quad I'_{F_f-I_f} = G \big( C(I_f, F_f) \big ).
%\label{gen-self}
%\end{equation}

%
Then, ($F_s$, $I_s$) and ($F_f$, $I_f$) are regarded as the inputs of the generator $G$ to respectively produce a pseudo shadow image $I'_{F_s, I_s}$ and a pseudo non-shadow image $I'_{F_f, I_f}$ as:
\begin{align}
        I'_{F_s, I_s} = G \big ( C(F_s, I_s) \big ), \\ I'_{F_f, I_f} = G \big( C(I_f, F_f) \big ).
\label{gen-self}
\end{align}

Due to the fact that the shadow feature and detail information are from the same image region identity, the generator $G$ should have the capacity to map $I'_{F_s, I_s}$ and $I'_{F_f, I_f}$ back to these regions of $I_s$ and $I_f$.
This is realized by the following reconstruction loss ${\cal L}_\text{rec}(\cdot, \cdot)$:
%
%
%Due to the shadow feature and details are come from the same region, the generator $G$ should be able to map $I'_{F_s-I_s}$ and $I'_{F_f-I_f}$ to the same as $I_s$, $I_f$, respectively.
%
%We apply the reconstruction loss ${\cal L}_\text{rec}(\cdot, \cdot)$ to achieve this goal, which is defined as:
%
\begin{equation}
       {\cal L}_\text{rec}(I_i, I'_{F_i, I_i}) = \| I_i - I'_{F_i, I_i} \|_1 + \ FFL( I_i - I'_{F_i, I_i}),
\label{rec-loss}
\end{equation}
where $\| \cdot \|_1$ denotes the $\ell_1$ loss. The $FFL(\cdot, \cdot)$ represents the focal frequency loss~\cite{jiang2021focal} that computes a weighted average of the distance between its two input images in the frequency domain. Narrowing the gap in the frequency domain has been demonstrated to improve the quality of synthesized images~\cite{jiang2021focal}.
Finally, our self-reconstruction constraint is defined as: 
\begin{equation}
        {\cal L}_\text{self}(E, G) = \omega_1 {\cal L}_\text{rec}(I_s, I'_{F_s, I_s}) + \omega_2 {\cal L}_\text{rec}(I_f, I'_{F_f, I_f}),
\label{self-loss}
\end{equation}
where $\omega_1, \omega_2$ are two trade-off parameters.

%The self-reconstruction loss gives the generator $G$ a overall perspective on what the pseudo images should be. The generator $G$ will be regularized to preserve the overall details of the image and avoid producing low-quality images.

%{\color{blue}This self-reconstruction process serves an important regularization role} to the learning of $G$~\cite{zheng2019joint} since {\color{blue}xxxxxxxxxxxxxxx}
%

\subsection{Inter-reconstruction}\label{inter-reconstruction}
Different from the self-reconstruction constraint that works with image synthesis using the same region identity for both encoder and generator, in this subsection, we focus on addressing the case of different region identities and an inter-reconstruction loss is further introduced.

According to Eq.\,(\ref{generate}), when $F_j = F_s$ and $I_i = I_f$, the generator will produce a pseudo shadow image $I'_{F_s,I_f}$ as:
\begin{equation}
        I'_{F_s, I_f} = G \big ( C(F_s, I_f) \big ),
\label{gen-inter}
\end{equation}
which, we expect, to possess a similar shadow to the shadow region $I_s$ and the same details in $I_f$.
Similarly, when $F_j = F_f$ and $I_i = I_s$, the generator $G$ will produce a pseudo non-shadow image $I'_{F_f, I_s}$ as:
\begin{equation}
    I'_{F_f, I_s} = G \big( C(F_f, I_s) \big ),
\end{equation}
which, we expect, to have a similar shadow to the non-shadow region $I_f$, and the same details in $I_s$.

To retain the shadow information, as shown in Fig.\,\ref{fig:framework}, we introduce a shadow discriminator $D_s$ and a non-shadow discriminator $D_f$ to adversarially train our HQSS. In our setting, the $I'_{F_s, I_f}$ tries to trick the $D_s$ while $I'_{F_f, I_s}$ tries to trick the $D_f$. Inspired by LSGAN~\cite{mao2017least}, we build the following objective to stabilize the network training:
\begin{equation}
    \begin{aligned}
        {\cal L}_\text{Dis}(E, G) & =  \frac{1}{2} \big( D_s(I'_{F_s, I_f}) - 1 \big)^2 \\& + \frac{1}{2} \big( D_f(I'_{F_f, I_s}) - 1 \big)^2.
    \end{aligned}
\label{G-Dloss}
\end{equation}

In the meantime, the $D_s$ and $D_f$ are expected to distinguish the pseudo shadow $I'_{F_s, I_f}$ from real shadow region $I_s$ and the pseudo non-shadow $I'_{F_f, I_s}$ from real non-shadow region $I_f$. Thus, the following regularization is further introduced:
\begin{align}
%    \begin{aligned}
        {\cal L}_\text{Dis}(D_s)  & = \big( D_s(I'_{F_s, I_f}) \big)^2  + \big( D_s(I_s) - 1 \big)^2,  \label{D-loss}\\
        {\cal L}_\text{Dis}(D_f)  & =  \big( D_f(I'_{F_f, I_s}) \big)^2  + \big( D_f(I_f) - 1 \big)^2. \label{D-loss2}
%    \end{aligned}
\end{align}

By optimizing Eq.\,(\ref{G-Dloss}) and Eq.\,(\ref{D-loss}) and Eq.\,(\ref{D-loss2}) in a min-max game manner, these two discriminators learn to distinguish the pseudo images and the real images, which in turn forces the shadow feature encoder $E$ to extract the correct shadow feature and the generator $G$ to learn to synthesize real-like and in-distribution shadow/non-shadow pseudo images.

\begin{figure}[!t]
\centering
\includegraphics[width=0.48\textwidth]{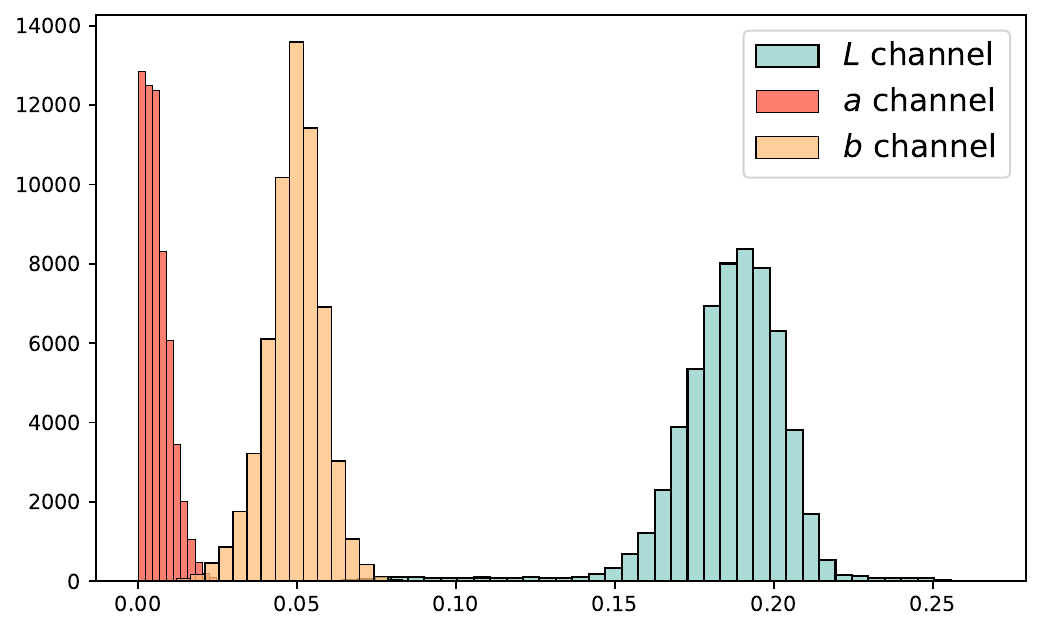}
\caption{Example of the absolute difference between shadow image and its non-shadow counterpart on L, A, B channel, respectively.}
\label{fig:color-diff}
\end{figure}

Nevertheless, the adversarial training simply derives pseudo images with shadow/non-shadow characteristics. It overlooks preserving detail information.
Fortunately, we observe that details are mostly preserved in $A$, $B$ channels if images are represented on LAB color space.
Fig.\,\ref{fig:color-diff} gives an illustrative example. In $L$ channel that reflects the lightness, a significant difference is observed between the shadow region and the non-shadow one, while in $A$, $B$ channels that reflect colors, the difference is relatively minor which means that these two channels preserve most image details.
Motivated by this finding, we further propose to minimize the distance in $A$, $B$ channels between the pseudo image $I'_{F_j - I_i}$ and the decoupled region $I_i$ to indirectly preserve details as:
%
%However, the adversarial training only encourages the pseudo image to be shadow or non-shadow, which overlooks the preservation of details. Fortunately, in LAB color space, a prior-based loss can be adapt to alleviate this problem. 
%
%An illustrative example is given in Fig.\,\ref{fig:color-diff}, in \textit{L} channel that reflects the lightness, the difference between the shadow region and its non-shadow version is larger which indicates the lightness changes. While in A, B channel that reflects colors, the difference is relatively minor which means that these two channels mostly preserve the details of the image.
%
%Thus, we design a color consistency loss:
\begin{equation}
    \begin{aligned}
        {\cal L}_\text{color}(E, G)   & =  \| \mathbb{T}_{ab}(I'_{F_s, I_f})- \mathbb{T}_{ab}(I_f)\|_1  \\
        &+ \| \mathbb{T}_{ab}(I'_{F_f, I_s}) - \mathbb{T}_{ab}(I_s) \|_1 ,
    \end{aligned}
\label{color-loss}
\end{equation}
where $\mathbb{T}_{ab}(\cdot)$ returns the A, B channels of the input.
%
%, $M_{I_f}, M_{I_s} \in \{0, 1\}$ are respectively the mask of $I_f, I_s$, $\odot$ denotes element-wise multiplication.
%
%The multiplication of mask is to emphasize the non-zero area of the $I_f$ and $I_s$.
%
%In line with a prior knowledge, the color consistency loss minimizes the distance between the pseudo image and its details image on A, B channel to preserve the details.

Combining Eq.\,(\ref{G-Dloss}), Eq.\,(\ref{D-loss}), Eq.\,(\ref{D-loss2}), and Eq.\,(\ref{color-loss}) forms our inter-reconstruction loss as: 
\begin{equation}
    \begin{aligned}
        {\cal L}_\text{inter}(E, G, D_s, D_f) & = \omega_3 {\cal L}_\text{Dis}(E, G) + \omega_4 {\cal L}_\text{color}(E, G) \\ & + \omega_5 {\cal L}_\text{Dis}(D_s) + \omega_6 {\cal L}_\text{Dis}(D_f),
    \end{aligned}
\label{inter-loss}
\end{equation}
where $\omega_3, \omega_4, \omega_5$ and $\omega_6$ are the trade-off parameters.
%
%Under the guidance of the inter-reconstruction loss, the shadow feature encoder $E$ learns to extract the shadow feature from the given region. 
%
%The generator $G$ learns to arrange the pseudo image to be a shadow/non-shadow image according to the input shadow feature and having the details as the input details image. 

\subsection{Cycle-reconstruction}\label{cycle-reconstruction}

Except for the aforementioned inter-reconstruction loss, we further introduce a cycle-reconstruction loss where the synthesized shadow/non-shadow images are in return considered as the inputs of our HQSS to ensure that they share the same shadow feature and detail information as the real shadow/non-shadow image regions.

Given the pseudo shadow image $I'_{F_s - I_f}$ and pseudo non-shadow image $I'_{F_f - I_s}$, their shadow features can be extracted by the encoder $E$ as:
%
%The third training objective is cycle-reconstruction to make sure that the pseudo image from inter-reconstruction should be able to provide the same shadow feature and the same details as that used in generating it.
%
%
%
%Firstly, to ensure the pseudo image has the same shadow feature as the reference image, the shadow feature of the pseudo image extracted by $E$ should be identical to that shadow feature used to generate it.
%
%In particular, given $I'_{F_s-I_f}$ and $I'_{F_f-I_s}$, their shadow feature can be extracted by the shadow feature encoder:
%
\begin{equation}
        F'_s = E(I'_{F_s, I_f}), \quad F'_f = E(I'_{F_f, I_s}).
\label{ext-E-cycle}
\end{equation}

From the encoder side, it is natural to expect the shadow features $F'_s$ and $F'_f$ of the synthesized pseudo images to align with $F_s$ and $F_f$ of these real images. This is realized by the following shadow feature cycle constraint:
%
%Since $I'_{F_s-I_f}$ and $I'_{F_f-I_s}$ are respectively generated with shadow feature $F_s$ and $F_f$, the extracted $F_s'$ and $F_f'$ are required to be same as them one by one with the following shadow feature cycle loss:
%
\begin{equation}
        {\cal L}_\text{shadow}(E, G) = \| F_s - F'_s \|_1 +  \| (F_f - F'_f) \|_1.
\label{sha-cycle-loss}
\end{equation}

%Due to $F_s$ and $F_f$ respectively coming from $I_s$ and $F_s$, the applying of the shadow feature cycle loss will make $I'_{F_s-I_f}$ has the same shadow as $I_s$, and $I'_{F_f-I_s}$ has the same shadow as $I_f$.
%
%Thus, the shadow feature cycle loss guarantees that the pseudo image captures the characteristics of the real shadow.

%
Then, we respectively pair the synthesized images $I'_{F_s, I_f}$ and $I'_{F_f, I_s}$ with the real shadow features $F_f$ and $F_s$ as inputs of the generator $G$ and derive the following outputs:
%
%Secondly, to ensure the pseudo image has the same details as the details image, the pseudo image should be able to provide enough details information when it is regarded as the details image.
%
%To this end, $I'_{F_s-I_f}$ is paired with shadow feature $F_f$ to input the generator $G$ to produce a pseudo image $I'_{F_f-I'_f}$, and $I'_{F_f-I_s}$ and $F_s$ to input the generator $G$ to produce $I'_{F_s-I'_s}$.
%
\begin{align}
        I'_{F_f, I'_f} = G \big ( C(F_f, I'_{F_s, I_f}) \big ), \\  I'_{F_s, I'_s} = G \big ( C(F_s, I'_{F_f, I_s})\big ).
\label{gen-cycle}
\end{align}

It is intuitive that $I'_{F_f, I'_f}$ and $I'_{F_s, I'_s}$ should be identical with $I_f$ and $I_s$, respectively.
We employ the reconstruction loss ${\cal L}_\text{rec}(\cdot, \cdot)$ of Eq.\,(\ref{rec-loss}) to realize this identical mapping:
\begin{equation}
        {\cal L}_\text{rec-cycle}(E, G) = {\cal L}_\text{rec}(I_f, I'_{F_f, I'_f}) + {\cal L}_\text{rec}(I_s, I'_{F_s, I'_s}).
\label{rec-cycle-loss}
\end{equation}

%Through minimizing Eq.\,(\ref{rec-cycle-loss}), $I'_{F_f-I'_f}$ and $I'_{F_s-I'_s}$ are required to possess the same details as $I_f$ and $I_s$, respectively. 

Combining Eq.\,(\ref{sha-cycle-loss}) and Eq.\,(\ref{rec-cycle-loss}) leads to our cycle-reconstruction loss as: 
\begin{equation}
    \begin{aligned}
        {\cal L}_\text{cycle}(E, G) & = \omega_7 {\cal L}_\text{rec-cycle}(E, G) + \omega_8 {\cal L}_\text{shadow}(E, G),
    \end{aligned}
\label{cycle-loss}
\end{equation}
where $\omega_7$ and $\omega_8$ are the trade-off hyper-parameters.

%These two terms in the cycle-reconstruction loss require the pseudo image from inter-reconstruction to equip the same shadows as these of the real images used to generate it and the same details as to its own details image. Thus, the authenticity of the pseudo shadow is guaranteed and the details information can be preserved as much as possible.

%These two terms in the cycle-reconstruction loss require the pseudo image from inter-reconstruction to equip the same shadow as that of the reference image used to generate it and the same details as to its own details image. Thus, the authenticity of the pseudo shadow is guaranteed and the details information can be preserved as much as possible.

\subsection{Shadow Removal}

Combining the proposed losses of self-reconstruction, inter-reconstruction, and cycle-reconstruction leads to the overall training objective:
%
%Integrating all of the proposed training objectives, the total training objective for the four sub-networks in our paper is defined as:
\begin{equation}
    \begin{aligned}
        {\cal L}_\text{total}(E, G, D_s, D_f) & = {\cal L}_\text{self}(E, G) + {\cal L}_\text{inter}(E, G, D_s, D_f) \\ &+ {\cal L}_\text{cycle}(E, G).
    \end{aligned}
\label{total-loss}
\end{equation}

As for the trade-off parameters of $\omega_1 \sim \omega_8$, their values are empirically set to balance the magnitude of each loss item as:
$\omega_1=1$, $\omega_2=1$, $\omega_3=0.05$, $\omega_4=0.01$, $\omega_5=1$, $\omega_6=1$, $\omega_7=0.1$, $\omega_8=0.01$.

%We set the value of all trade-off hyper-parameters in Eq.\,(\ref{total-loss}) to make sure the magnitude of each loss is balanced. Thus, the configurations of these trade-off hyper-parameters are $\omega_1=1$, $\omega_2=1$, $\omega_3=0.05$, $\omega_4=0.01$, $\omega_5=1$, $\omega_6=1$, $\omega_7=0.1$, $\omega_8=0.01$.

After the training of our HQSS, $I'_{F_s - I_f}$ and $I'_{F_f - I_s}$ embrace the shadow characteristic as well as the detail information.
Then, $I'_{F_s - I_f}$ and $I_f$ form paired shadow/non-shadow data to train any existing shadow removal networks. 
For a fair comparison with the recent state-of-the-art, we take the shadow removal network employed in G2R-ShadowNet~\cite{liu2021shadowCVPR} as an illustrative example to show the efficacy of our synthesized images in this paper.
%
%After training, the generator $G$ and the shadow feature encoder $E$ take a reference image $I_s$ and a details image $I_f$ to produce a pseudo shadow image $I'_{F_s-I_f}$. Then $I'_{F_s-I_f}$ and its non-shadow version $I_f$ form the paired data to train the following shadow removal network. For a fair comparison, we employ the shadow removal network in G2R-ShadowNet~\cite{liu2021shadowCVPR}. 

%Specifically, the shadow removal network consists of two sub-networks including an inverse network $N_{iv}$ and a refinement network $N_r$.
%
%The inverse network $N_{iv}$ takes $I'_{F_s-I_f}$ as input to produce a preliminary result $R_i$:
%
Briefly speaking, the shadow removal network of G2R-ShadowNet consists of an inverse network $N_{iv}$ and a refinement network $N_r$. In our setting, the inverse network $N_{iv}$ takes the pseudo shadow image $I'_{F_s, I_f}$ as its input to produce a non-shadow result $R_i$:
\begin{equation}
    \begin{aligned}
        R_i = N_{iv}(I'_{F_s, I_f}).
    \end{aligned}
\label{pre-result}
\end{equation}
%

%With output $R_i$, the inverse network $N_{iv}$ is trained with:
%
Then, $R_i$ is paired with $I_f$ to form an $\ell_1$ loss for training the inverse network $N_{iv}$:
\begin{equation}
    \begin{aligned}
        {\cal L}_\text{inverse}(N_{iv}) = \| R_i, I_f\|_1.
    \end{aligned}
\label{inverse-loss}
\end{equation}

To train the refinement network $N_r$, $I'_{F_s, I_f}$ is firstly added to the original shadow image $I$ by:
%
%$R_i$ is then embedded with the input whole shadow image $I$ by:
%
\begin{equation}
    \begin{aligned}
        R_e = \big (R_i \otimes M' \big ) \oplus  \big (I \otimes (1 - M') \big ),
    \end{aligned}
\label{embed}
\end{equation}
where $M' \in \{0, 1\}$ is the shadow mask of $I'_{F_s, I_f}$, $\otimes$ denotes element-wise multiplication, and $\oplus$ means element-wise addition. The output $R_e$ is fed to the refinement network $N_r$ to produce a refined result $R_f$:
\begin{equation}
    \begin{aligned}
        R_f = N_r(R_e),
    \end{aligned}
\label{fin-result}
\end{equation}
which is guided by the $\ell_1$ training loss as:
%
%And the refinement loss is defined as:
%
\begin{equation}
    \begin{aligned}
        {\cal L}_\text{refine}(N_{iv}, N_{r}) = \| R_f, I\|_1 + \| M_d \odot R_f, M_d \odot I \|_1,
    \end{aligned}
\label{ref-loss}
\end{equation}
%where the $M_d$ denotes a dilation mask with $50$ kernel size. The second term pays more attention to the adjacent area of the shadow part. Note the computed gradient will be propagated back both $N_{iv}$ and $N_r$. The total loss for shadow removal network consists of Eq.\,(\ref{inverse-loss}) and Eq.\,(\ref{ref-loss}):
%
where the $M_d$ denotes a dilation mask with $50$ kernel size. The second term pays more attention to the adjacent area of the shadow region.

Combining Eq.\,(\ref{inverse-loss}) and Eq.\,(\ref{ref-loss}) results in the overall training objective for the shadow removal network of G2R-ShadowNet as:

\begin{equation}
    \begin{aligned}
        {\cal L}_\text{removal}(N_{iv}, N_{r}) = {\cal L}_\text{inverse} + {\cal L}_\text{refine}.
    \end{aligned}
\label{total-removal-loss}
\end{equation}
%
%where the $\gamma_1=1, \gamma_2=1$.

\section{Experiments}

\subsection{Implementation Details}

%We implement all code using  Pytorch~\cite{paszke2019pytorch}. 
%
%The shadow feature encoder $E$ consists of one convolutional layer and two residual blocks that each have two convolutional layers. The resolution is kept the same as the input.
%
%The generator $G$ consists is basically the same as that in~\cite{hu2019mask,liu2021shadowCVPR}. It contains one convolutional layer with a stride of 1, two convolutional layers with a stride of 2 to lower the resolution, nine residual blocks that each has two convolutional layers, followed by two deconvolutional layers with a stride of 2 to restore the resolution, and ends with a convolutional layer to produce the final output.
%
%After each convolutional layer, the instance normalization~\cite{ulyanov2016instance} and a ReLU function are applied.
%
%The discriminators $D_s$ and $D_f$ are the same as PatchGAN~\cite{isola2017image}. It consists of three convolutional layers with a stride of 2, two convolutional layers with a stride of 1, and ends with an average pooling layer.
%
%The inverse network $N_{iv}$ and the refinement network $N_r$ have the same architecture as the generator $G$.

In our implementation, the shadow feature encoder $E$ consists of one convolutional layer and two residual blocks constituted by two convolutional layers. As for the generator $G$, following~\cite{hu2019mask,liu2021shadowCVPR}, it contains one convolutional layer with a stride of 1, two convolutional layers with a stride of 2 to lower the resolution, nine residual blocks that each has two convolutional layers, followed by two deconvolutional layers with a stride of 2 to restore the resolution, and ends with a convolutional layer to produce the final output. Besides, After each convolutional layer, the instance normalization~\cite{ulyanov2016instance} and a ReLU function are inserted.
As for the discriminators $D_s$ and $D_f$, we directly borrow from PatchGAN~\cite{isola2017image} in which each discriminator consists of three convolutional layers with a stride of 2, two convolutional layers with a stride of 1, and ends with an average pooling layer.
Lastly, the inverse network $N_{iv}$ and the refinement network $N_r$ in the shadow removal network share the same architecture with the generator $G$.

To train the proposed HQSS and the shadow removal network, the batch size is set to 1, and Adam optimizer~\cite{kingma2014adam} is adopted with the first momentum of 0.9 and the second momentum of 0.999.
The learning rate is initialized to 2$\times$10$^{-4}$ and is linearly decayed to $0$ in the last 50 epochs.
%
%The initial learning rate is set to $2 \times 10^{-4}$ for the first $E-50$ epochs and it will be linear decay to $0$ for the rest $50$ epochs. The $E$ denotes the total training epochs.
%
We train the proposed HQSS for 100 epochs and the shadow removal network for 150 epochs.
All network is initialized using a Gaussian distribution with a mean of 0 and a standard deviation of 0.02.

All the input images and output images are represented in the LAB color space. 
Each input image is first resized to 448$\times$448 and then randomly cropped to a 400$\times$400 image. We also apply random flipping to enhance the generalization.
We use the same way as in G2R-ShadowNet~\cite{liu2021shadowCVPR} to generate the random non-shadow mask that is used to crop a non-shadow region identity from the input image $I$.
Specifically, we randomly select a shadow mask from the mask set of other training images. The selected mask is then applied to the non-shadow part of $I$ to obtain the non-shadow region identity.
All experiments are implemented using Pytorch~\cite{paszke2019pytorch}. 

\begin{table*}
\begin{center}
\caption{Quantitative results on ISTD dataset. ``Shadow Region'' indicates the metrics are computed on the shadow part of the testing images, while ``Non-Shadow Region '' indicates the non-shadow part.
``RMSE$^*$'' denotes that the RMSE is calculated by averaging the RMSE of all pixels. 
``RMSE'' denotes that the RMSE is calculated by averaging the RMSE over all pixels in each image and then averaging the sum RMSE of test images.
Results of other methods are obtained from \cite{liu2021shadowCVPR} or the original paper.}
\label{comparsion:ISTD}
\begin{tabular}{c|c|cccc|ccc|ccc}
\toprule[1.25pt]
\multirow{2}{*}{Methods} & \multirow{2}{*}{Data} & \multicolumn{4}{c|}{Shadow Region} & \multicolumn{3}{c|}{Non-Shadow Region} & \multicolumn{3}{c}{Whole Image} \\ \cline{3-12} 
                         &                       &  RMSE$^*$ & RMSE   & PSNR  & SSIM   &  RMSE     & PSNR      &   SSIM    &  RMSE  &  PSNR     &  SSIM    \\ \midrule[0.75pt]
     Input Image    &        -              & 39.01  & 36.95  & 20.83   & 0.927 &  2.42       &   37.46     &    0.985  &  8.40   &   20.46  &  0.894 \\ \midrule[0.5pt]
Yang \emph{et al.}~\cite{yang2012shadow}&      -           &  24.7  &   23.2 & 21.57  & 0.878   &  14.2     &   22.25  &  0.782   &    15.9   &  20.26  & 0.706  \\
Gong \emph{et al.}~\cite{gong2014interactive}&          -             &  13.3 & 13.0   & 30.53   & 0.972  &    2.6   &   36.63    &  0.982  &  4.3  &  28.96   &  0.943  \\ \midrule[0.5pt]
Guo \emph{et al.}~\cite{guo2012paired}  & Paired+Mask & 22.0  &  20.1  &  26.89 & 0.960 &   3.1    &   35.48   &   0.975    & 6.1  &   25.51 &  0.924 \\ ST-CGAN~\cite{wang2018stacked}
 &  Paired+Mask &  13.4  &  12.0 &  31.70  & 0.979  &   7.9    &   26.39    &  0.956  &   8.6  &  24.75   &  0.927  \\ 
SP+M-Net~\cite{le2019shadow} &   Paired+Mask &  7.9   &  8.1  &  35.08  & 0.984 &    2.8    &   36.38    &   0.979  & 3.6    & 31.89 &  0.953 \\
 G2R-ShadowNet (\emph{Sup})~\cite{liu2021shadowCVPR} &  Paired+Mask& 7.3 &  7.9   & 36.12    & 0.988  & 2.9   & 35.21   &  0.977   & 3.6  & 31.93   &    0.957           \\\midrule[0.5pt]
 Mask-ShadowGAN~\cite{hu2019mask}   & Unpaired Images  & 9.9   & 10.8   &  32.19 &  0.984   &  3.8  &  33.44  &   0.974    &  4.8    &  28.81   &   0.946  \\
 LG-ShadowNet~\cite{liu2021shadow}    &  Unpaired Images  & 9.7   & 9.9    &   32.44  & 0.982 & 3.4        &  33.68      &  0.971   &   4.4   & 29.20 & 0.945 \\ 
 DC-ShadowNet~\cite{jin2021dc} &  Unpaired Images  &  -  &  10.3   &   -  & - & 3.5       &  -      &  -   &   4.6  & - & - \\ \midrule[0.5pt]
 FSS2SR~\cite{le2020shadow}   &    Shadow+Mask      & 9.7    &  10.4   &  33.09  & \textbf{0.983}  &    2.9     &   35.26  &   0.977    &  4.0    &  30.12    &  \textbf{0.950}   \\
  G2R-ShadowNet~\cite{liu2021shadowCVPR}      &       Shadow+Mask       & 8.8   &  8.9   &    33.58 &        0.979    &   2.9    &    35.52  &  0.976  &  3.9   & 30.52 &  0.944  \\
 HQSS (Ours)      &      Shadow+Mask       & \textbf{8.27}   &  \textbf{8.48}   & \textbf{33.94}   &  0.980  &  \textbf{2.82}        &  \textbf{35.59}      &    \textbf{0.978}    & \textbf{3.72}   &   \textbf{30.76}   &  0.948   \\ \bottomrule[0.75pt]
\end{tabular}
\end{center}
\end{table*}

\subsection{Datasets and Metrics}

\subsubsection{Datasets} We conduct experiments on three widely-used datasets including the ISTD dataset, the Video Shadow Removal dataset, and the SRD dataset.

%ISTD dataset~\cite{wang2018stacked} consists of 1,870 triplets of shadow, shadow mask, and shadow-free images where 1,330 triplets are used for training the rest of 540 triplets are used for testing. In our paper, we employ the adjusted testing set~\cite{le2019shadow} where the color inconsistency between shadow and non-shadow images in the original dataset is reduced. 
%
ISTD dataset~\cite{wang2018stacked} consists of 1,870 triplets of (shadow, shadow mask, shadow-free image). Among them, 1,330 triplets are used for training and the remaining are used for testing. In this paper, we employ the adjusted testing set~\cite{le2019shadow} where the color inconsistency phenomenon between shadow and non-shadow images is well reduced. 
In the training stage of the proposed HQSS and the shadow removal network, we use the shadow images and ground-truth shadow masks from the training set.
In the inference stage of the shadow removal network, the shadow masks of the test set are the shadow detector~\cite{zhu2018bidirectional}, which is pre-trained on the training sets of ISTD and SBU~\cite{vicente2016large} and it achieves 2.4 Balance Error Rate on the testing set of the ISTD dataset.

Video Shadow Removal dataset~\cite{le2020shadow} contains 8 videos in a static scene.
A total of 997 images are extracted from these 8 videos for testing.
It provides a $V_{max}$ image for each video as the non-shadow ground truth and a simple formula that sets a threshold to obtain the moving-shadow mask for each video.
Though the moving-shadow masks are a coarse estimation of the ground truth mask, we use them to evaluate our method following previous work~\cite{le2020shadow}.
We train our model on the ISTD dataset and directly test it on Video Shadow Removal Dataset. 
In the inference stage of the shadow removal network, we utilize the shadow mask generated by the shadow detector~\cite{zhu2018bidirectional} pre-trained only on SBU dataset as~\cite{liu2021shadowCVPR}.

SRD dataset~\cite{qu2017deshadownet} has 2,680 and 408 images for training and testing.
Due to the lack of shadow mask in the original SRD dataset, Cun \emph{et al}.~\cite{cun2020towards} have provided a coarse estimation.
We use these masks to train and evaluate our proposed HQSS and the shadow removal network.
In the inference stage, the same masks are used.

\subsubsection{Evaluation metrics}
Following common practice~\cite{wang2018stacked,hu2019mask}, we use the Root-Mean-Square Error (RMSE), Peak Signal-to-Noise Ratio (PSNR), and Structural Similarity (SSIM) to evaluate the performance of the proposed HQSS. 
The RMSE is computed between the predicted non-shadow result and its ground-truth counterpart in LAB color space.
We provide two types of RMSE including one that is directly averaged over all the pixels and another that is averaged among pixels in each image first and then averaged over all images. %As stated in~\cite{liu2021shadowCVPR}, the latter RMSE pay reflects more the quality of each image on shadow and non-shadow parts. 
The PSNR and SSIM scores are computed in the RGB color space.
A lower RMSE indicates a better result, and a higher PSNR and SSIM denote better performance. 
For the ISTD dataset and Video Shadow Removal dataset, the output is resized to 256$\times$256 for evaluation. As for the SRD dataset, the output is resized to $640 \times 840$ for evaluation.

\subsection{Performance Comparison}

\begin{figure*}[!t]
\centering
\includegraphics[width=0.95\textwidth]{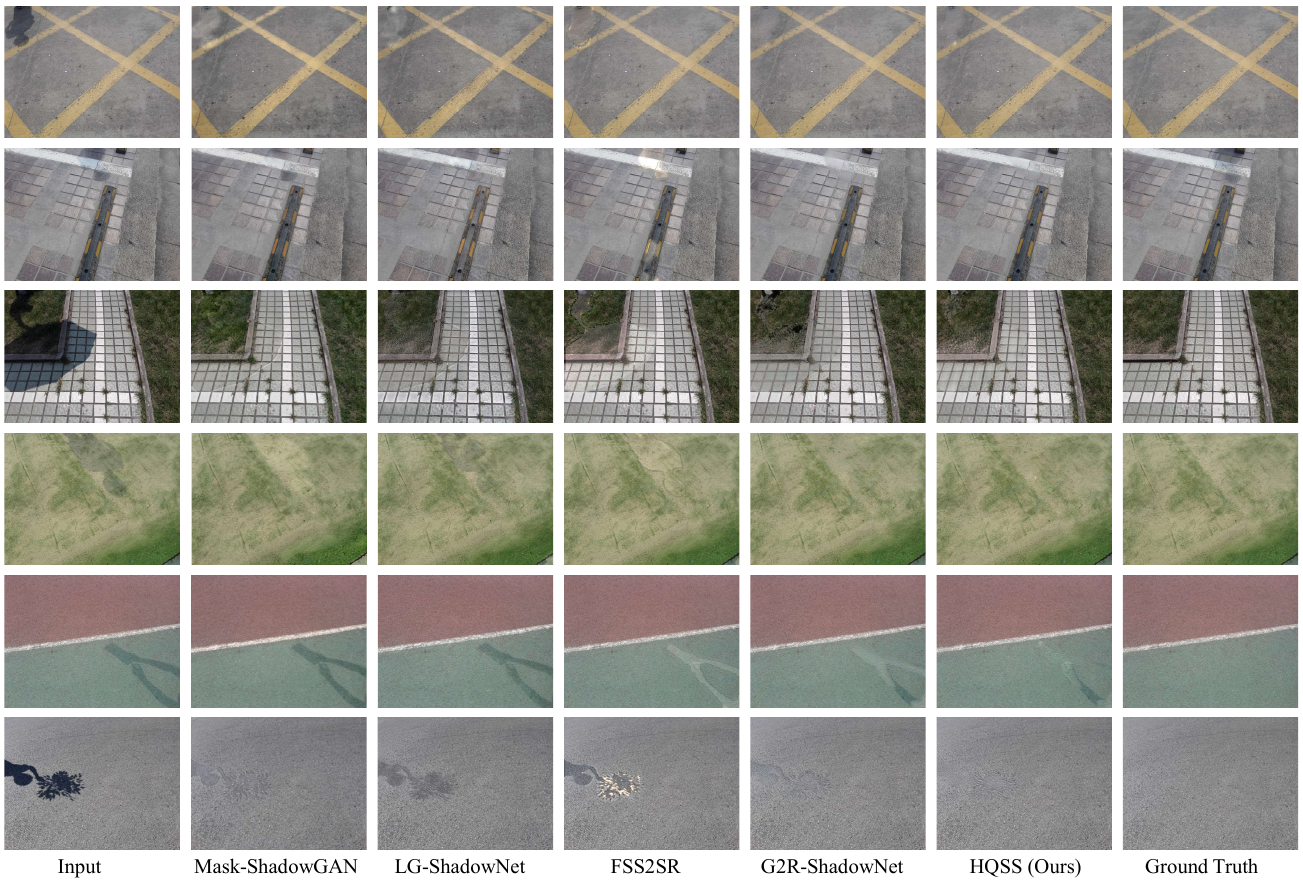}
\caption{Visualization comparisons between other state-of-the-art methods and the proposed HQSS. Samples come from the ISTD dataset.}
\label{fig:vis}
\end{figure*}

\begin{table}
\begin{center}
\caption{Quantitative results on Video Shadow Removal dataset. 
We provide the results on the moving-shadow region.
%
%The metrics are calculated on the moving-shadow region. 
%
RMSE$_{40}$ denotes that the moving-shadow mask is computed with a threshold of 40. Other results use the moving-shadow mask which is computed with a threshold of 80.
All results of other methods are obtained from the paper of \cite{liu2021shadowCVPR}.}
\label{comparsion:video}
\begin{tabular}{c|cccc}
\toprule[1.25pt]
Methods               & RMSE                 & RMSE$_{40}$                 & PSNR                 & SSIM                 \\ \midrule[0.75pt]
SP+M-Net~\cite{le2019shadow}  &          -         &   22.2          &    -              &   -               \\
Mask-ShadowGAN~\cite{hu2019mask}&      22.7            &       19.6            &        20.38       &   0.887              \\
LG-ShadowNet~\cite{liu2021shadow} &  22.0               &   18.3             &      20.68            &    0.880             \\
FSS2SR~\cite{le2020shadow}  &          -         &    20.9        &     -           &       -            \\
G2R-ShadowNet~\cite{liu2021shadowCVPR} &     21.8            &     18.8             &        21.07        & 0.882                  \\
HQSS (Ours) & \textbf{18.95} & \textbf{16.82} &  \textbf{21.89} &  \textbf{0.888}\\ \bottomrule[0.75pt]
\end{tabular}
\end{center}
\end{table}

\begin{table*}
\begin{center}
\caption{Quantitative results on SRD dataset. ``\dag'' indicates the results are reproduced via using their official code.
``\ddag'' indicates the results are from our re-implementation for the unavailability of public code.}
\label{comparsion:srd}
\begin{tabular}{c|c|cccc|ccc|ccc}
\toprule[1.25pt]
\multirow{2}{*}{Methods} & \multirow{2}{*}{Data} & \multicolumn{4}{c|}{Shadow Region} & \multicolumn{3}{c|}{Non-Shadow Region} & \multicolumn{3}{c}{Whole Image} \\ \cline{3-12} 
                         &                       &  RMSE$^*$ & RMSE   & PSNR  & SSIM   &  RMSE     & PSNR      &   SSIM    &  RMSE  &  PSNR     &  SSIM    \\ \midrule[0.75pt]
     Input Image    &        -              & 42.09  & 46.39  & 19.54   & 0.875 &  4.92       &   30.26     &    0.972  &  13.83   &   18.26  &  0.837 \\ \midrule[0.5pt]
 Mask-ShadowGAN\dag~\cite{hu2019mask}   & Unpaired Images  & 11.62   & 11.02   &  31.16 &  0.964   &  6.07  &  28.39  &   0.911    &  7.11    &  25.78   &   0.868  \\
 LG-ShadowNet\dag~\cite{liu2021shadow}    &  Unpaired Images  & 17.15   &  13.85  & 29.98  &  0.952 & 6.42        &  28.24      &  0.918   &   8.49   & 24.99 & 0.864 \\ \midrule[0.5pt]
 FSS2SR\ddag~\cite{le2020shadow}   &    Shadow+Mask      & 39.02 & 38.43 & 20.90& 0.878 & 15.03 &20.21 & 0.873 & 20.19 &16.08 & 0.743 \\
  G2R-ShadowNet\dag~\cite{liu2021shadowCVPR}      &       Shadow+Mask       & 25.74   &  22.14   &    26.45 &   0.921    &   5.84    &    27.86   &  0.942  &  9.86  & 23.04 &  0.849  \\
 HQSS (Ours)      &      Shadow+Mask       & \textbf{18.70}   &  \textbf{18.05}   & \textbf{27.03}   & \textbf{0.946}    &    \textbf{5.84}      &   \textbf{27.86}     &  \textbf{0.942}      &  \textbf{8.36}  &   \textbf{23.71}   &  \textbf{0.878}    \\ \bottomrule[0.75pt]
\end{tabular}
\end{center}
\end{table*}

\subsubsection{Results on ISTD dataset} 
%
%in this subsection, we provide comparisons between the proposed HQSS and several state-of-the-art methods on ISTD dataset. The quantitative results are shown in Tab.\,\ref{comparsion:ISTD}. 
%
%The approach of \cite{yang2012shadow} and \cite{gong2014interactive} use the image priors to remove shadows and the training data is not involved.
%
%\cite{guo2012paired,wang2018stacked,le2019shadow}, and G2R-ShadowNet (\emph{Sup}) utilize the paired shadow/non-shadow images and shadow mask to perform the shadow removal task. Note that we adopt the same shadow removal network as G2R-ShadowNet, thus the results of G2R-ShadowNet (\emph{Sup}) are also the performance of training our shadow removal network with paired data in a supervised manner.
%
%Mask-ShadowGAN~\cite{hu2019mask} and
%LG-ShadowNet~\cite{liu2021shadow} both require unpaired shadow and non-shadow images for training.
%
%FSS2SR~\cite{le2020shadow} and G2R-ShadowNet~\cite{liu2021shadowCVPR} are the same type as ours where only the shadow images and shadow masks are required.
%
Tab.\,\ref{comparsion:ISTD} displays the quantitative comparison with many existing methods~\cite{yang2012shadow,gong2014interactive,guo2012paired,wang2018stacked,le2019shadow,hu2019mask,liu2021shadow,le2020shadow,liu2021shadowCVPR} on ISTD dataset.
\cite{yang2012shadow,gong2014interactive} consider the image priors to remove shadows without the involvement of training data.
\cite{guo2012paired,wang2018stacked,le2019shadow} and G2R-ShadowNet (\emph{Sup}) utilize the paired shadow/non-shadow images and shadow mask to perform the shadow removal task. Note that we adopt the same shadow removal network as G2R-ShadowNet, thus the results of G2R-ShadowNet (\emph{Sup}) are also the performance of training our shadow removal network with paired data in a supervised manner.
Mask-ShadowGAN~\cite{hu2019mask} and
LG-ShadowNet~\cite{liu2021shadow} both require unpaired shadow and non-shadow images for training.
Similar to our method, FSS2SR~\cite{le2020shadow} and G2R-ShadowNet~\cite{liu2021shadowCVPR} only require shadow images and shadow masks.

As the first row block shows, the input image itself leads to inferior results in the shadow region.
When compared with the results from the second row block, the proposed HQSS significantly outperforms the two prior-based methods. It is worth emphasizing that~\cite{gong2014interactive} is an interactive method where the shadow and non-shadow part is obtained with the assistance of human.
The methods in the third row block use paired data and shadow mask to train their shadow removal networks.
Despite only relying on the shadow image and the shadow mask, our method achieves competitive performance compared with them. For example, we obtain 8.27 RMSE$^*$ on the shadow region which outperforms Guo \emph{et al.}~\cite{guo2012paired} and ST-CGAN~\cite{wang2018stacked}. 
In comparison to LG-ShadowNet~\cite{liu2021shadow} and Mask-ShadowGAN~\cite{hu2019mask}, our method respectively obtains $1.63$ and $1.43$ improvement on RMSE$^*$ even though the non-shadow images are not involved in our method.
A fair comparison would be that with  FSS2SR~\cite{le2020shadow} and G2R-ShadowNet~\cite{liu2021shadowCVPR} as they only use the shadow images and shadow masks.
We can observe that our method consistently outperforms FSS2SR on RMSE and PSNR for all shadow regions, non-shadow regions, and the whole image. Only two of the SSIM values are marginally lower than their results.
For the comparison with G2R-ShadowNet, our method obtains a better performance for all metrics in each region.

Fig.\,\ref{fig:vis} provides the visualization results. It can be observed that the non-shadow images of our method are more realistic where the color and image texture are better preserved compared with other methods. These visualization comparisons are in line with the quantitative results.

\subsubsection{Results on Video Shadow Removal dataset} 
We continue testing the performance on the Video Shadow Removal dataset while training existing methods on the ISTD dataset.
%
%
%in this subsection, we conduct the comparisons between our method and other state-of-the-art approaches on Video Shadow Removal dataset. The methods in this subsection are trained on ISTD dataset and then directly tested on Video Shadow Removal dataset. 
Tab.\,\ref{comparsion:video} provides the quantitative results.
SP+M-Net~\cite{le2019shadow} uses the paired data and shadow mask in the training process. Mask-ShadowGAN~\cite{hu2019mask} and LG-ShadowNet~\cite{liu2021shadow} need unpaired data for training. FSS2SR~\cite{le2020shadow}, G2R-ShadowNet~\cite{liu2021shadowCVPR}, and ours only require the shadow images and shadow masks.
It can be seen that the proposed HQSS achieves performance improvements for all metrics, in particular to RMSE and RMSE$_{40}$. These results demonstrate that our HQSS possesses better generalization ability than other methods.

\subsubsection{Results on SRD dataset} We report the quantitative comparisons with other state-of-the-art approaches on the challenging SRD dataset in Tab.\,\ref{comparsion:srd}. 

Compared with methods~\cite{liu2021shadow,hu2019mask} in the second row block that consider the non-shadow images in the training process, our method achieves better performance when tested on the non-shadow region.
As for the comparison with FSS2SR~\cite{le2020shadow} and G2R-ShadowNet~\cite{liu2021shadowCVPR} that adopt the same data type as ours, the proposed HQSS outperforms them by a large margin when tested on the shadow region and the whole image.
The significant improvement occurs in RMSE$^*$ of the shadow region where our method outperforms G2R-ShadowNet by 7.04.

\begin{table}
\begin{center}
\caption{Ablation study of each component in the proposed HQSS. ``-'' indicates removing the corresponding component.}
\label{ablation}
\begin{tabular}{l|ccc|ccc}
\toprule[1.25pt]
\multirow{2}{*}{Methods} & \multicolumn{3}{c|}{Shadow Region}                                  & \multicolumn{3}{c}{Whole Image}                                            \\ \cline{2-7} 
                         & RMSE                 & PSNR                 & SSIM                  & \multicolumn{1}{c}{RMSE} & \multicolumn{1}{c}{PSNR} & SSIM                 \\ \midrule[0.75pt]
   $-$ ${\cal L}_\text{self}$  & 8.98 &   33.31    &  0.979    &   3.84      &   30.32      &    0.946   \\
   $-$ ${\cal L}_\text{cycle}$  & 9.59 &  33.13  &   0.978   &  4.01                      &     30.08                 &    0.943              \\
  $-$ ${\cal L}_\text{self}/{\cal L}_\text{cycle}$  &   9.61         &     32.95         &     0.975              &        4.01       &    30.02                 &      0.941           \\
\midrule[0.75pt]
$-$ ${\cal L}_\text{color}$ &   9.03     &    33.79          &  0.979           &             3.85              &  30.70   &      0.947           \\
     $-$ $I'_{F_f,I_s}$   &   9.15            &      33.54           &  0.979                &     3.94                 &     30.37           &   0.946             \\
  $-$ ${\cal L}_\text{Dis}$   &   35.12              &        21.58        &  0.925          &    8.13                 &      21.86              &    0.890          
 \\\midrule[0.75pt]
  All (Ours) & \textbf{8.48} & \textbf{33.94} & \textbf{0.980} &    \textbf{3.72}         &   \textbf{30.76}                &  \textbf{0.948} \\ \bottomrule[0.75pt]
\end{tabular}
\end{center}
\end{table}

\subsection{Ablation Study}\label{Ablation Study}

In this section, we conduct experiments to demonstrate the effectiveness of each component in our method. The quantitative results are reported in Tab.\,\ref{ablation}.

In this first row block of Tab.\,\ref{ablation}, we provide the quantitative results of removing self-reconstruction ${\cal L}_\text{self}$ and cycle-reconstruction ${\cal L}_\text{cycle}$. Note that the inter-reconstruction has to be removed since the shadow removal network requires the pseudo shadow image produced by inter-reconstruction for training.
As the results show, the performance drops significantly when self-reconstruction ${\cal L}_\text{self}$ is removed, which in turn proves its efficacy.
When removing cycle-reconstruction ${\cal L}_\text{cycle}$, we can see the performance also drops by a large margin.
Thus, the capacity of cycle-reconstruction ${\cal L}_\text{cycle}$ in preserving authenticity and details information of the pseudo shadow is inevitable. 
When the self-reconstruction ${\cal L}_\text{self}$ and cycle-reconstruction ${\cal L}_\text{cycle}$ are simultaneously removed, the performance degrades severely.

In this second row block of Tab.\,\ref{ablation}, we provide the ablation studies about the removal of the components in the inter-reconstruction. We can observe that the removal of the color consistency loss incurs large performance degradation, which indicates this prior-based loss benefits preserving detail information. 
When removing all losses that are relevant to $I'_{F_f, I_s}$, more performance degradation occurs. Thus, simply applying the proposed HQSS to learn with only the shadow part is insufficient. By regarding the non-shadow region as a special shadow, the shadow feature encoder $E$ and the generator $G$ are adequately trained via learning how to reconstruct both shadow/non-shadow parts, which leads to higher performance.
The discriminators in our method are also necessary. Otherwise, the model training will collapse.

\section{Conclusion}

In this paper, we investigate the problem of synthesizing high-quality pseudo shadow images for the shadow removal task by only using the shadow images and corresponding shadow masks.
We introduce a novel shadow generation framework HQSS which employs a shadow feature encoder and a generator to synthesize high-quality pseudo images.
A given image is first decoupled into a shadow region identity and a non-shadow region identity.
The shadow feature encoder is used to extract the shadow feature of a region identity. Then the extracted shadow feature is paired with another region identity to form the input of the generator to synthesize a pseudo image.
We expect the pseudo image to have the shadow feature as its input shadow feature and a similar image detail as its input region identity.
To this end, we design three learning objectives.
A self-reconstruction loss is proposed to guide the generator to reconstruct an identical pseudo image as its input if the shadow feature and input region identity are from the same region identity.
An inter-reconstruction loss and a cycle-reconstruction loss are introduced to guide the generator to retain shadow characteristics and detail information if the shadow feature and input region identity are from different identities
Based on the proposed three learning objectives, our HQSS is able to synthesize high-quality pseudo shadow images. Correspondingly, the shadow removal network trained on these high-quality pseudo shadow images achieves better performance than other state-of-the-art methods, which clearly demonstrates the effectiveness of our HQSS.

\bibliographystyle{IEEEtran}
\bibliography{egbib}

\begin{thebibliography}{10}
\providecommand{\url}[1]{#1}
\csname url@samestyle\endcsname
\providecommand{\newblock}{\relax}
\providecommand{\bibinfo}[2]{#2}
\providecommand{\BIBentrySTDinterwordspacing}{\spaceskip=0pt\relax}
\providecommand{\BIBentryALTinterwordstretchfactor}{4}
\providecommand{\BIBentryALTinterwordspacing}{\spaceskip=\fontdimen2\font plus
\BIBentryALTinterwordstretchfactor\fontdimen3\font minus
  \fontdimen4\font\relax}
\providecommand{\BIBforeignlanguage}[2]{{%
\expandafter\ifx\csname l@#1\endcsname\relax
\typeout{** WARNING: IEEEtran.bst: No hyphenation pattern has been}%
\typeout{** loaded for the language `#1'. Using the pattern for}%
\typeout{** the default language instead.}%
\else
\language=\csname l@#1\endcsname
\fi
#2}}
\providecommand{\BIBdecl}{\relax}
\BIBdecl

\bibitem{le2019shadow}
H.~Le and D.~Samaras, ``Shadow removal via shadow image decomposition,'' in
  \emph{Proceedings of the IEEE/CVF International Conference on Computer Vision
  (ICCV)}, 2019, pp. 8578--8587.

\bibitem{nie2021unsupervised}
L.~Nie, C.~Lin, K.~Liao, S.~Liu, and Y.~Zhao, ``Unsupervised deep image
  stitching: Reconstructing stitched features to images,'' \emph{IEEE
  Transactions on Image Processing (TIP)}, vol.~30, pp. 6184--6197, 2021.

\bibitem{zhang2020content}
J.~Zhang, C.~Wang, S.~Liu, L.~Jia, N.~Ye, J.~Wang, J.~Zhou, and J.~Sun,
  ``Content-aware unsupervised deep homography estimation,'' in
  \emph{Proceedings of the European Conference on Computer Vision
  (ECCV)}.\hskip 1em plus 0.5em minus 0.4em\relax Springer, 2020, pp. 653--669.

\bibitem{shen2020ransac}
X.~Shen, F.~Darmon, A.~A. Efros, and M.~Aubry, ``Ransac-flow: generic two-stage
  image alignment,'' in \emph{Proceedings of the European conference on
  computer vision (ECCV)}.\hskip 1em plus 0.5em minus 0.4em\relax Springer,
  2020, pp. 618--637.

\bibitem{wang2008texture}
Z.-Z. Wang and J.-H. Yong, ``Texture analysis and classification with linear
  regression model based on wavelet transform,'' \emph{IEEE transactions on
  image processing (TIP)}, vol.~17, no.~8, pp. 1421--1430, 2008.

\bibitem{yu2018generative}
J.~Yu, Z.~Lin, J.~Yang, X.~Shen, X.~Lu, and T.~S. Huang, ``Generative image
  inpainting with contextual attention,'' in \emph{Proceedings of the IEEE/CVF
  Conference on Computer Vision and Pattern Recognition (CVPR)}, 2018, pp.
  5505--5514.

\bibitem{liu2018image}
G.~Liu, F.~A. Reda, K.~J. Shih, T.-C. Wang, A.~Tao, and B.~Catanzaro, ``Image
  inpainting for irregular holes using partial convolutions,'' in
  \emph{Proceedings of the European conference on computer vision (ECCV)},
  2018, pp. 85--100.

\bibitem{yeh2017semantic}
R.~A. Yeh, C.~Chen, T.~Yian~Lim, A.~G. Schwing, M.~Hasegawa-Johnson, and M.~N.
  Do, ``Semantic image inpainting with deep generative models,'' in
  \emph{Proceedings of the IEEE/CVF Conference on Computer Vision and Pattern
  Recognition (CVPR)}, 2017, pp. 5485--5493.

\bibitem{Zamir2020MIRNet}
S.~W. Zamir, A.~Arora, S.~Khan, M.~Hayat, F.~S. Khan, M.-H. Yang, and L.~Shao,
  ``Learning enriched features for real image restoration and enhancement,'' in
  \emph{Proceedings of the European Conference on Computer Vision (ECCV)},
  2020.

\bibitem{zamir2021multi}
S.~W. Zamir, A.~Arora, S.~Khan, M.~Hayatx, F.~S. Khan, M.-H. Yang, and L.~Shao,
  ``Multi-stage progressive image restoration,'' in \emph{Proceedings of the
  IEEE/CVF Conference on Computer Vision and Pattern Recognition (CVPR)}, 2021,
  pp. 14\,821--14\,831.

\bibitem{zamir2022restormer}
S.~W. Zamir, A.~Arora, S.~Khan, M.~Hayat, F.~S. Khan, and M.-H. Yang,
  ``Restormer: Efficient transformer for high-resolution image restoration,''
  in \emph{Proceedings of the IEEE/CVF Conference on Computer Vision and
  Pattern Recognition (CVPR)}, 2022, pp. 5728--5739.

\bibitem{tao2018scale}
X.~Tao, H.~Gao, X.~Shen, J.~Wang, and J.~Jia, ``Scale-recurrent network for
  deep image deblurring,'' in \emph{Proceedings of the IEEE/CVF Conference on
  Computer Vision and Pattern Recognition (CVPR)}, 2018, pp. 8174--8182.

\bibitem{zhang2019deep}
H.~Zhang, Y.~Dai, H.~Li, and P.~Koniusz, ``Deep stacked hierarchical
  multi-patch network for image deblurring,'' in \emph{Proceedings of the
  IEEE/CVF Conference on Computer Vision and Pattern Recognition (CVPR)}, 2019,
  pp. 5978--5986.

\bibitem{huang2011characterizes}
X.~Huang, G.~Hua, J.~Tumblin, and L.~Williams, ``What characterizes a shadow
  boundary under the sun and sky?'' in \emph{Proceedings of the IEEE/CVF
  International Conference on Computer Vision (ICCV)}, 2011, pp. 898--905.

\bibitem{finlayson2005removal}
G.~D. Finlayson, S.~D. Hordley, C.~Lu, and M.~S. Drew, ``On the removal of
  shadows from images,'' \emph{IEEE transactions on pattern analysis and
  machine intelligence (TPAMI)}, vol.~28, no.~1, pp. 59--68, 2005.

\bibitem{guo2012paired}
R.~Guo, Q.~Dai, and D.~Hoiem, ``Paired regions for shadow detection and
  removal,'' \emph{IEEE transactions on pattern analysis and machine
  intelligence (TPAMI)}, vol.~35, no.~12, pp. 2956--2967, 2012.

\bibitem{khan2015automatic}
S.~H. Khan, M.~Bennamoun, F.~Sohel, and R.~Togneri, ``Automatic shadow
  detection and removal from a single image,'' \emph{IEEE transactions on
  pattern analysis and machine intelligence (TPAMI)}, vol.~38, no.~3, pp.
  431--446, 2015.

\bibitem{zhang2015shadow}
L.~Zhang, Q.~Zhang, and C.~Xiao, ``Shadow remover: Image shadow removal based
  on illumination recovering optimization,'' \emph{IEEE Transactions on Image
  Processing (TIP)}, vol.~24, no.~11, pp. 4623--4636, 2015.

\bibitem{yang2012shadow}
Q.~Yang, K.-H. Tan, and N.~Ahuja, ``Shadow removal using bilateral filtering,''
  \emph{IEEE Transactions on Image processing (TIP)}, vol.~21, no.~10, pp.
  4361--4368, 2012.

\bibitem{wang2018stacked}
J.~Wang, X.~Li, and J.~Yang, ``Stacked conditional generative adversarial
  networks for jointly learning shadow detection and shadow removal,'' in
  \emph{Proceedings of the IEEE/CVF Conference on Computer Vision and Pattern
  Recognition (CVPR)}, 2018, pp. 1788--1797.

\bibitem{le2020shadow}
H.~Le and D.~Samaras, ``From shadow segmentation to shadow removal,'' in
  \emph{Proceedings of the European Conference on Computer Vision
  (ECCV)}.\hskip 1em plus 0.5em minus 0.4em\relax Springer, 2020, pp. 264--281.

\bibitem{hu2019mask}
X.~Hu, Y.~Jiang, C.-W. Fu, and P.-A. Heng, ``Mask-shadowgan: Learning to remove
  shadows from unpaired data,'' in \emph{Proceedings of the IEEE/CVF
  International Conference on Computer Vision (ICCV)}, 2019, pp. 2472--2481.

\bibitem{liu2021shadowCVPR}
Z.~Liu, H.~Yin, X.~Wu, Z.~Wu, Y.~Mi, and S.~Wang, ``From shadow generation to
  shadow removal,'' in \emph{Proceedings of the IEEE/CVF Conference on Computer
  Vision and Pattern Recognition (CVPR)}, 2021, pp. 4927--4936.

\bibitem{liu2021shadow}
Z.~Liu, H.~Yin, Y.~Mi, M.~Pu, and S.~Wang, ``Shadow removal by a
  lightness-guided network with training on unpaired data,'' \emph{IEEE
  Transactions on Image Processing (TIP)}, vol.~30, pp. 1853--1865, 2021.

\bibitem{gao2022towards}
J.~Gao, Q.~Zheng, and Y.~Guo, ``Towards real-world shadow removal with a shadow
  simulation method and a two-stage framework,'' in \emph{Proceedings of the
  IEEE/CVF Conference on Computer Vision and Pattern Recognition Workshops
  (CVPRW)}, 2022, pp. 599--608.

\bibitem{li2019asymmetric}
Y.~Li, S.~Tang, R.~Zhang, Y.~Zhang, J.~Li, and S.~Yan, ``Asymmetric gan for
  unpaired image-to-image translation,'' \emph{IEEE Transactions on Image
  processing (TIP)}, vol.~28, no.~12, pp. 5881--5896, 2019.

\bibitem{choi2018stargan}
Y.~Choi, M.~Choi, M.~Kim, J.-W. Ha, S.~Kim, and J.~Choo, ``Stargan: Unified
  generative adversarial networks for multi-domain image-to-image
  translation,'' in \emph{Proceedings of the IEEE/CVF Conference on Computer
  Vision and Pattern Recognition (CVPR)}, 2018, pp. 8789--8797.

\bibitem{zhu2017unpaired}
J.-Y. Zhu, T.~Park, P.~Isola, and A.~A. Efros, ``Unpaired image-to-image
  translation using cycle-consistent adversarial networks,'' in
  \emph{Proceedings of the IEEE/CVF International Conference on Computer Vision
  (ICCV)}, 2017, pp. 2223--2232.

\bibitem{gryka2015learning}
M.~Gryka, M.~Terry, and G.~J. Brostow, ``Learning to remove soft shadows,''
  \emph{ACM Transactions on Graphics (TOG)}, vol.~34, no.~5, pp. 1--15, 2015.

\bibitem{shor2008shadow}
Y.~Shor and D.~Lischinski, ``The shadow meets the mask: Pyramid-based shadow
  removal,'' in \emph{Computer Graphics Forum}, vol.~27, no.~2.\hskip 1em plus
  0.5em minus 0.4em\relax Wiley Online Library, 2008, pp. 577--586.

\bibitem{xiao2013fast}
C.~Xiao, R.~She, D.~Xiao, and K.-L. Ma, ``Fast shadow removal using adaptive
  multi-scale illumination transfer,'' in \emph{Computer Graphics Forum},
  vol.~32, no.~8.\hskip 1em plus 0.5em minus 0.4em\relax Wiley Online Library,
  2013, pp. 207--218.

\bibitem{zhang2018improving}
W.~Zhang, X.~Zhao, J.-M. Morvan, and L.~Chen, ``Improving shadow suppression
  for illumination robust face recognition,'' \emph{IEEE transactions on
  pattern analysis and machine intelligence (TPAMI)}, vol.~41, no.~3, pp.
  611--624, 2018.

\bibitem{liu2008texture}
F.~Liu and M.~Gleicher, ``Texture-consistent shadow removal,'' in
  \emph{Proceedings of the European Conference on Computer Vision
  (ECCV)}.\hskip 1em plus 0.5em minus 0.4em\relax Springer, 2008.

\bibitem{vicente2017leave}
T.~F.~Y. Vicente, M.~Hoai, and D.~Samaras, ``Leave-one-out kernel optimization
  for shadow detection and removal,'' \emph{IEEE Transactions on Pattern
  Analysis and Machine Intelligence (TPAMI)}, vol.~40, no.~3, pp. 682--695,
  2017.

\bibitem{fredembach2005hamiltonian}
C.~Fredembach and G.~Finlayson, ``Hamiltonian path-based shadow removal,'' in
  \emph{British Machine Vision Conference (BMVC)}, vol.~2, no. CONF, 2005, pp.
  502--511.

\bibitem{ding2019argan}
B.~Ding, C.~Long, L.~Zhang, and C.~Xiao, ``Argan: Attentive recurrent
  generative adversarial network for shadow detection and removal,'' in
  \emph{Proceedings of the IEEE/CVF International Conference on Computer Vision
  (ICCV)}, 2019, pp. 10\,213--10\,222.

\bibitem{qu2017deshadownet}
L.~Qu, J.~Tian, S.~He, Y.~Tang, and R.~W. Lau, ``Deshadownet: A multi-context
  embedding deep network for shadow removal,'' in \emph{Proceedings of the
  IEEE/CVF Conference on Computer Vision and Pattern Recognition (CVPR)}, 2017,
  pp. 4067--4075.

\bibitem{hu2019direction}
X.~Hu, C.-W. Fu, L.~Zhu, J.~Qin, and P.-A. Heng, ``Direction-aware spatial
  context features for shadow detection and removal,'' \emph{IEEE transactions
  on pattern analysis and machine intelligence (TPAMI)}, vol.~42, no.~11, pp.
  2795--2808, 2019.

\bibitem{zhang2020ris}
L.~Zhang, C.~Long, X.~Zhang, and C.~Xiao, ``Ris-gan: Explore residual and
  illumination with generative adversarial networks for shadow removal,'' in
  \emph{Proceedings of the AAAI Conference on Artificial Intelligence (AAAI)},
  vol.~34, no.~07, 2020, pp. 12\,829--12\,836.

\bibitem{cun2020towards}
X.~Cun, C.-M. Pun, and C.~Shi, ``Towards ghost-free shadow removal via dual
  hierarchical aggregation network and shadow matting gan,'' in
  \emph{Proceedings of the AAAI Conference on Artificial Intelligence (AAAI)},
  vol.~34, no.~07, 2020, pp. 10\,680--10\,687.

\bibitem{fu2021auto}
L.~Fu, C.~Zhou, Q.~Guo, F.~Juefei-Xu, H.~Yu, W.~Feng, Y.~Liu, and S.~Wang,
  ``Auto-exposure fusion for single-image shadow removal,'' in
  \emph{Proceedings of the IEEE/CVF Conference on Computer Vision and Pattern
  Recognition (CVPR)}, 2021, pp. 10\,571--10\,580.

\bibitem{chen2021canet}
Z.~Chen, C.~Long, L.~Zhang, and C.~Xiao, ``Canet: A context-aware network for
  shadow removal,'' in \emph{Proceedings of the IEEE/CVF International
  Conference on Computer Vision (ICCV)}, 2021, pp. 4743--4752.

\bibitem{jin2021dc}
Y.~Jin, A.~Sharma, and R.~T. Tan, ``Dc-shadownet: Single-image hard and soft
  shadow removal using unsupervised domain-classifier guided network,'' in
  \emph{Proceedings of the IEEE/CVF International Conference on Computer Vision
  (ICCV)}, 2021, pp. 5027--5036.

\bibitem{sato2003illumination}
I.~Sato, Y.~Sato, and K.~Ikeuchi, ``Illumination from shadows,'' \emph{IEEE
  Transactions on Pattern Analysis and Machine Intelligence (TPAMI)}, vol.~25,
  no.~3, pp. 290--300, 2003.

\bibitem{arief2012realtime}
I.~Arief, S.~McCallum, and J.~Y. Hardeberg, ``Realtime estimation of
  illumination direction for augmented reality on mobile devices,'' in
  \emph{Color and Imaging Conference}, vol. 2012, no.~1.\hskip 1em plus 0.5em
  minus 0.4em\relax Society for Imaging Science and Technology, 2012, pp.
  111--116.

\bibitem{karsch2011rendering}
K.~Karsch, V.~Hedau, D.~Forsyth, and D.~Hoiem, ``Rendering synthetic objects
  into legacy photographs,'' \emph{ACM Transactions on Graphics (TOG)},
  vol.~30, no.~6, pp. 1--12, 2011.

\bibitem{karsch2014automatic}
K.~Karsch, K.~Sunkavalli, S.~Hadap, N.~Carr, H.~Jin, R.~Fonte, M.~Sittig, and
  D.~Forsyth, ``Automatic scene inference for 3d object compositing,''
  \emph{ACM Transactions on Graphics (TOG)}, vol.~33, no.~3, pp. 1--15, 2014.

\bibitem{gardner2019deep}
M.-A. Gardner, Y.~Hold-Geoffroy, K.~Sunkavalli, C.~Gagn{\'e}, and J.-F.
  Lalonde, ``Deep parametric indoor lighting estimation,'' in \emph{Proceedings
  of the IEEE/CVF International Conference on Computer Vision (ICCV)}, 2019,
  pp. 7175--7183.

\bibitem{zhang2019all}
J.~Zhang, K.~Sunkavalli, Y.~Hold-Geoffroy, S.~Hadap, J.~Eisenman, and J.-F.
  Lalonde, ``All-weather deep outdoor lighting estimation,'' in
  \emph{Proceedings of the IEEE/CVF Conference on Computer Vision and Pattern
  Recognition (CVPR)}, 2019, pp. 10\,158--10\,166.

\bibitem{hold2019deep}
Y.~Hold-Geoffroy, A.~Athawale, and J.-F. Lalonde, ``Deep sky modeling for
  single image outdoor lighting estimation,'' in \emph{Proceedings of the
  IEEE/CVF Conference on Computer Vision and Pattern Recognition (CVPR)}, 2019,
  pp. 6927--6935.

\bibitem{goodfellow2014generative}
I.~J. Goodfellow, J.~Pouget-Abadie, M.~Mirza, B.~Xu, D.~Warde-Farley, S.~Ozair,
  A.~Courville, and Y.~Bengio, ``Generative adversarial nets,'' in
  \emph{Proceedings of the Advances in Neural Information Processing Systems
  (NeurIPS)}, 2014, pp. 2672--2680.

\bibitem{zhang2019shadowgan}
S.~Zhang, R.~Liang, and M.~Wang, ``Shadowgan: Shadow synthesis for virtual
  objects with conditional adversarial networks,'' \emph{Computational Visual
  Media}, vol.~5, no.~1, pp. 105--115, 2019.

\bibitem{liu2020arshadowgan}
D.~Liu, C.~Long, H.~Zhang, H.~Yu, X.~Dong, and C.~Xiao, ``Arshadowgan: Shadow
  generative adversarial network for augmented reality in single light
  scenes,'' in \emph{Proceedings of the IEEE/CVF Conference on Computer Vision
  and Pattern Recognition (CVPR)}, 2020, pp. 8139--8148.

\bibitem{jiang2021focal}
L.~Jiang, B.~Dai, W.~Wu, and C.~C. Loy, ``Focal frequency loss for image
  reconstruction and synthesis,'' in \emph{Proceedings of the IEEE/CVF
  International Conference on Computer Vision (ICCV)}, 2021, pp.
  13\,919--13\,929.

\bibitem{mao2017least}
X.~Mao, Q.~Li, H.~Xie, R.~Y. Lau, Z.~Wang, and S.~Paul~Smolley, ``Least squares
  generative adversarial networks,'' in \emph{Proceedings of the IEEE/CVF
  International Conference on Computer Vision (ICCV)}, 2017, pp. 2794--2802.

\bibitem{ulyanov2016instance}
D.~Ulyanov, A.~Vedaldi, and V.~Lempitsky, ``Instance normalization: The missing
  ingredient for fast stylization,'' \emph{arXiv preprint arXiv:1607.08022},
  2016.

\bibitem{isola2017image}
P.~Isola, J.-Y. Zhu, T.~Zhou, and A.~A. Efros, ``Image-to-image translation
  with conditional adversarial networks,'' in \emph{Proceedings of the IEEE/CVF
  Conference on Computer Vision and Pattern Recognition (CVPR)}, 2017, pp.
  1125--1134.

\bibitem{kingma2014adam}
D.~P. Kingma and J.~Ba, ``Adam: A method for stochastic optimization,'' in
  \emph{Proceedings of the International Conference on Learning Representations
  (ICLR)}, 2014.

\bibitem{paszke2019pytorch}
A.~Paszke, S.~Gross, F.~Massa, A.~Lerer, J.~Bradbury, G.~Chanan, T.~Killeen,
  Z.~Lin, N.~Gimelshein, L.~Antiga \emph{et~al.}, ``Pytorch: An imperative
  style, high-performance deep learning library,'' in \emph{Proceedings of the
  Advances in Neural Information Processing Systems (NeurIPS)}, 2019, pp.
  8026--8037.

\bibitem{gong2014interactive}
H.~Gong and D.~Cosker, ``Interactive shadow removal and ground truth for
  variable scene categories,'' in \emph{British Machine Vision Conference
  (BMVC)}.\hskip 1em plus 0.5em minus 0.4em\relax BMVA Press, 2014.

\bibitem{zhu2018bidirectional}
L.~Zhu, Z.~Deng, X.~Hu, C.-W. Fu, X.~Xu, J.~Qin, and P.-A. Heng,
  ``Bidirectional feature pyramid network with recurrent attention residual
  modules for shadow detection,'' in \emph{Proceedings of the European
  Conference on Computer Vision (ECCV)}, 2018, pp. 121--136.

\bibitem{vicente2016large}
T.~F.~Y. Vicente, L.~Hou, C.-P. Yu, M.~Hoai, and D.~Samaras, ``Large-scale
  training of shadow detectors with noisily-annotated shadow examples,'' in
  \emph{Proceedings of the European Conference on Computer Vision
  (ECCV)}.\hskip 1em plus 0.5em minus 0.4em\relax Springer, 2016, pp. 816--832.

\end{thebibliography}

\end{document}